# Tracking object's type changes with fuzzy based fusion rule

Albena Tchamova, Jean Dezert and Florentin Smarandache

**Abstract**

In this paper the behavior of three combinational rules for temporal/sequential attribute data fusion for target type estimation are analyzed. The comparative analysis is based on: Dempster's fusion rule proposed in Dempster-Shafer Theory; Proportional Conflict Redistribution rule no. 5 (PCR5), proposed in Dezert-Smarandache Theory and one alternative class fusion rule, connecting the combination rules for information fusion with particular fuzzy operators, focusing on the t-norm based Conjunctive rule as an analog of the ordinary conjunctive rule and t-conorm based Disjunctive rule as an analog of the ordinary disjunctive rule. The way how different t-conorms and t-norms functions within TCN fusion rule influence over target type estimation performance is studied and estimated.

## 1 Introduction

Target type estimates can be used during different target tracking process' stages for improving data to track association and for the quality evaluation of complicated situations characterized with closely spaced or/and crossing targets. It supports the process of identification, (e.g. friendly aircraft against hostile ones, fighter against cargo), helping that way the operator of the tracking system to initiate relevant measures. The process of identification could use many types of different attribute information, for instance, discrete information associated with ESM (Electronic Support Measures) data, IRST (Infra-Red Search and Track) measurements and direct target type observations. It requires also efficient fusion rules and criteria to estimate the correct associations. Our motivation for attribute fusion is inspired from the necessity to ascertain the targets' types, information, that in consequence has an important implication to enhance the tracking performance. In the literature we can find many fusion methods derived from different underlying frameworks, ranging from Bayesian probability theory ([1]), Dempster-Shafer evidence theory ([2], [3], [4]) to fuzzy sets ([5], [6], [7]) etc. In temporal multiple target tracking the main requirements we have to deal with relate especially to the way of adequate conflict processing/redistribution, the simplicity of implementation, satisfaction of the impact of neutrality of Vacuous Belief Assignment (VBA), reflection of majority opinion, etc. The choice of method ([15]) depends on the richness of abstraction and diversity of sensor data.

The most used until now Dempster-Shafer Theory (DST) proposes a suitable mathematical framework for representation of uncertainty. It is one of the widely applied framework in the area of target tracking when one wants to deal with uncertain information and take into account attribute data and/or human-based information. DST, thanks to belief functions, is well suited for combining the information obtained by independent sources (bodies of evidence), especially in case of low conflicts between them. Although very appealing, DST presents some weaknesses and limitations and can give rise to some paradoxes/anomalies ([8]), [13], [14]), failed to provide the correct solution for some specific association problems, when the conflict increases and becomes very high (close to 1). To overcome the drawbacks of Dempster's fusion rule and in the meantime to extend the domain of application of the belief functions, recently a new mathematical framework, called Dezert-Smarandache Theory (DSmT) ([8], [9], [10]) was proposed for solving fusion/association problems, overcoming the practical limitations of DST, coming essentially from its inherent constraints, which are closely related with the acceptance of the law of the third excluded middle and can be interpreted as a general and direct extension of probability theory and the DST. The basic idea of DSmT is to work on Dedekind's lattice (called Hyper-Power Set) rather than on the classical power set of the frame as proposed in DST and, when needed, DSmT can also take into account the integrity constraints on elements of the frame, constraints which can also sometimes change over time with new knowledge. Hence DSmT deals with uncertain, imprecise and high conflicting information for static and dynamic fusion as well. Recently there is defined a new set of combination rules within DSmT, among them the Proportional Conflict Redistribution no. 5 ([9], [11]) which proposes a sophisticated and efficient solution for information fusion even when Shafer's model is considered and the conflict between the sources of information increase. In this paper we focus our attention on the ability of one alternative class fusion rule, connecting the combination rules for information fusion with particular fuzzy operators, focusing on the t-norm based conjunctive rule as an analog of the ordinary conjunctive rule of combination and on t-conorm based disjunctive rule as an analog of the ordinary disjunctive rule of combination, the so called T-Conorm-Norm (TCN) fusion rule ([12]). It is defined within DSmT and based on PCR5 fusion rule. The goal of fuzzy set/logic, which the TCN rule is based on, is to extend the classical binary logic into interval valued logic, thinking about something between false and truth. In this paper we concentrate our attention on the particular logical operator: *and*, implemented by the class of operators called t-norms and on the particular logical operator: *or*, implemented by the class of operators called t-conorms. We will study the way how different t-conorms and t-norms operators within TCN fusion rule influence over target type estimation performance. In the next sections we present briefly the basics of DSmT based PCR5 and TCN fusion rules, general properties and different types of t-conorm and t-norm functions. Then we describe the problem of object type estimation and identification. We apply the proposed TCN fusion rule for solving the problem and will evaluate how different types of t-norm functions influence over the estimation process, applied on a simple example. Then a comparative analysis of object type identification, obtained by using: *(1) Dempster's ([10]), PCR5 ([10]) and TCN fusion rules; (2) different types of t-conorm and t-norm functions within TCN fusion rule*; is provided on the base of estimated belief assignments. Concluding remarks are given in section 6.

## 2   Dezert-Smarandache theory based Proportional Conflict Redistribution rule no.5

Instead of distributing equally the total conflicting mass onto elements of power set as within Dempster's rule through the normalization step, or transferring the partial conflicts onto partial uncertainties as within DSm hybrid rule, the idea behind the Proportional Conflict Redistribution rules is to transfer conflicting masses (total or partial) proportionally to non-empty sets involved in the model according to all integrity constraints. The general principle is to :

- calculate the conjunctive rule of the belief masses of sources;

- calculate the total or partial conflicting masses ;

- redistribute the conflicting mass (total or partial) proportionally on non-empty sets involved in the model according to all integrity constraints.

These rules work both in DST and DSmT frameworks and for static or dynamical fusion problematic, for any degree of conflict in [0, 1], for any DSm models (Shafer's model, free DSm model or any hybrid DSm model). The most sophisticated rule among them is PCR5. For only two sources of information it is defined by:

$$m_{PCR5}(\varnothing) = 0$$

And for $\forall X \in G^{\Theta} \setminus \{\varnothing\}$,

$$m_{PCR5}(X) = m_{12}(X) + \sum_{\substack{Y \in G^{\Theta} \setminus \{X\} \\ c(X \cap Y) = \varnothing}} \left[ \frac{m_1(X)^2 m_2(Y)}{m_1(X) + m_2(Y)} + \frac{m_2(X)^2 m_1(Y)}{m_2(X) + m_1(Y)} \right] \quad (1)$$

Where $G^{\Theta}$ is the DSmT hyper-power set; $m_{12}(X)$ corresponds to the conjunctive consensus on $X$ between the two sources and where all denominators are different from zero and $c(X)$ is the canonical form of $X$ ([9], [11]).

No matter how big or small is the conflicting mass, PCR5 mathematically does a better redistribution of the conflicting mass than Dempster's rule and other rules since PCR5 goes backwards on the tracks of the conjunctive rule and redistributes the partial conflicting masses only to the sets involved in the conflict and proportionally to their masses put in the conflict, considering the conjunctive normal form of the partial conflict. PCR5 is quasi-associative and preserves the neutral impact of the vacuous belief assignment.

## 3   Fusion Rule based on Fuzzy T-Conorm-Norm Operators

The T-Conorm-Norm rule of combination represents a new class of fusion rules based on specified fuzzy t-Conorm, t-Norm operators. Triangular norms (t-norms) and Triangular conorms (t-conorms) are the most general families of binary functions that satisfy the requirements of the conjunction and disjunction operators, respectively. They are twoplace functions that map the unit square into the unit interval, i.e. $T-norm(x,y):[0,1]\times[0,1]\rightarrow[0,1]$ and $T-conorm(x,y):[0,1]\times[0,1]\rightarrow[0,1]$. They are monotonic, commutative and associative. TCN rule takes its source from the t-norm and t-conorm operators in fuzzy logics, where the *and* logic operator corresponds in information fusion to the conjunctive rule and the *or* logic operator corresponds to the disjunctive rule. In this work we propose to interpret the fusion/association between the sources of information as a vague relation, characterized with the following two characteristics:

- *The way of association between the possible propositions, built on the base of the frame of discernment.* It is based on union and intersection operations, and their combinations. These sets' operations correspond to logic operations conjunction and disjunction and their combinations.

- *The degree of association between the propositions.* It is obtained as a t-norm (for conjunction) or t-conorm (for disjunction) operators applied over the probability masses of corresponding focal elements. While the logic operators deal with degrees of truth and false, the fusion rules deal with degrees of belief of hypotheses. We will study the way how different t-norms functions within TCN fusion rule influence over target type estimation performance.

TCN fusion rule does not belong to the general Weighted Operator Class. The base principle of this rule developed in [12] consists in the following steps:

**Step 1**: Defining the T-norm conjunctive consensus:

The t-norm conjunctive consensus is based on the particular t-norm function. In general it is a function defined in fuzzy set/logic theory in order to represent the intersection between two particular fuzzy sets. If one extends t-norm to the data fusion theory, it will be a substitute for the conjunctive rule.

The t-norm has to satisfy the following conditions:

- Associativity:
$$T_{norm}(T_{norm}(x,y),z) = T_{norm}(x, T_{norm}(y,z)) \qquad (2)$$
- Commutativity:
$$T_{norm}(x,y) = T_{norm}(y,x) \qquad (3)$$
- Monotonicity:
$$if\ (x \leq a)\ \&\ (y \leq b)\ then\ T_{norm}(x,y) \leq T_{norm}(a,b) \qquad (4)$$
- Boundary Conditions:
$$T_{norm}(0,0) = 0;\ T_{norm(x,1)} = x \qquad (5)$$

There are many functions which satisfy these t-norm conditions:

- Zadeh's (default) min operator:

$$m(X) = \min\{m_1(X_i), m_2(X_j)\} \tag{6}$$

- Algebraic product operator:
$$m(X) = m_1(X_i) \cdot m_2(X_j) \tag{7}$$

- Bounded product operator:
$$m(X) = \max\{0, [m_1(X_i) + m_2(X_j) - 1]\} \tag{8}$$

The way of association between the focal elements of the given two sources of information, $m_1(.)$ and $m_2(.)$, is defined as $X = \theta_i \cap \theta_j$ and the degree of association is as follows:

$$\tilde{m}_{12}(X) = T_{norm}\{m_1(\theta_i), m_2(\theta_j)\} \tag{9}$$

where $\tilde{m}_{12}(X)$ represents the basic belief assignments (bba) after the fusion, associated with the given proposition $X$ by using particular t-norm based conjunctive rule.

***Step 2***: Distribution of the mass, assigned to the conflict.

In some degree it follows the distribution of conflicting mass in the most sophisticated DSmT based Proportional Conflict Redistribution rule number 5, but the procedure here relies on fuzzy operators. The total conflicting mass is distributed to all non-empty sets proportionally with respect to the *Maximum (Sum)* between the elements of corresponding mass matrix's columns, associated with the given element $X$ of the power set. It means the bigger mass is redistributed towards the element, involved in the conflict and contributing to the conflict with the maximum specified probability mass. The general procedure for fuzzy based conflict redistribution is as follows:

- Calculate all partial conflict masses separately;
- If $\theta_i \cap \theta_j = \varnothing$, then $\theta_i$ and $\theta_j$ are involved in the conflict; redistribute the corresponding masses $m_{12}(\theta_i \cap \theta_j) > 0$, involved in the particular partial conflicts to the non-empty sets $\theta_i$ and $\theta_j$ with respect to $\max\{m_1(\theta_i), m_2(\theta_j)\}$ and with respect to $\max\{m_1(\theta_j), m_2(\theta_i)\}$;
- Finally, for the given above two sources, the $T_{norm}$ conjunctive consensus yields:

$$\begin{aligned}
\tilde{m}_{12}(\theta_i) &= T_{norm}(m_1(\theta_i), m_2(\theta_j)) & \tilde{m}_{12}(\theta_j) &= T_{norm}(m_1(\theta_j), m_2(\theta_i)) \\
&+ T_{norm}(m_1(\theta_i), m_2(\theta_i \cup \theta_j)) & &+ T_{norm}(m_1(\theta_j), m_2(\theta_i \cup \theta_j)) \\
&+ T_{norm}(m_1(\theta_i \cup \theta_j), m_2(\theta_i)) & &+ T_{norm}(m_1(\theta_i \cup \theta_j), m_2(\theta_j))
\end{aligned} \tag{10}$$

$$\tilde{m}_{12}(\theta_i \cup \theta_j) = T_{norm}(m_1(\theta_i \cup \theta_j), m_2(\theta_i \cup \theta_j))$$

***Step 3***: The basic belief assignment, obtained as a result of the applied TCN rule becomes:

$$\tilde{m}_{TCN}(\theta_i) = \tilde{m}_{12}(\theta_i) + m_1(\theta_i) \times \frac{T_{norm}(m_1(\theta_i), m_2(\theta_j))}{T_{conorm}(m_1(\theta_i), m_2(\theta_j))}$$
$$+ m_2(\theta_i) \times \frac{T_{norm}(m_1(\theta_j), m_2(\theta_i))}{T_{conorm}(m_1(\theta_j), m_2(\theta_i))} ;$$

(11)

$$\tilde{m}_{TCN}(\theta_j) = \tilde{m}_{12}(\theta_j) + m_2(\theta_j) \times \frac{T_{norm}(m_1(\theta_i), m_2(\theta_j))}{T_{conorm}(m_1(\theta_i), m_2(\theta_j))}$$
$$+ m_1(\theta_i) \times \frac{T_{norm}(m_1(\theta_j), m_2(\theta_i))}{T_{conorm}(m_1(\theta_j), m_2(\theta_i))}$$

***Step 4:*** Normalization of the result.

The final step of the TCN fusion rule concerns the normalization procedure:

$$\tilde{m}_{TCN}(\theta) = \frac{\tilde{m}_{TCN}(\theta)}{\sum_{\substack{\theta \in 2^\Theta \\ \theta \neq \emptyset}} \tilde{m}_{TCN}(\theta)} \tag{12}$$

# 4 Object's type tracking issue

## 4.1 Formulation of the problem

The Target Type Tracking Problem can be simply stated as follows:

- Let $k = 1,...,k_{max}$ be the time index and consider $M$ possible target types: $T_i \in \Theta = \{\theta_1,...,\theta_M\}$, for example $\Theta = \{Fighter, Cargo\}$ with $T_1 \Rightarrow Fighter$, $T_2 \Rightarrow Cargo$, or $\Theta = \{Friend, Hostile\}$, etc.

- At each instant $k$, a target of true type $T(k) \in \Theta$ (not necessarily the same target) is observed by an attribute-sensor (we assume a perfect target detection probability here).

- the attribute measurement of the sensor (for example noisy Radar Cross Section) is then processed through a classifier which provides a decision $T_d(k)$ on the type of the observed target at each instant $k$.

- the sensor is in general not totally reliable and is characterized by a $M \times M$ confusion matrix:

$$C = [c_{ij} = P(T_d = T_j / TrueTargetType = T_i)] \tag{13}$$

Our goal is to estimate $T(k) \in \Theta$ from the sequence of declarations done by the unreliable classifier up to time $k$, i.e. to build an estimator:

$$\hat{T}(k = f(T_d(1), T_d(2),...,T_d(k))) \text{ of } T(k).\qquad(14)$$

## 4.2 Estimator's principle

In general the principle of our estimator is based on:

- a sequential temporal combination of the current basic belief assignment (drawn from classifier decision, i.e. our *measurements*) with the prior bba estimated up to current time from all past classifier declarations;

- Shafer's model for the frame of target types;

- One and the same information (vacuous belief assignment) as prior belief and same sequence of measurements (classifier declarations).

All the above assumptions are to get a fair comparative analysis according to: *(i)* performance evaluation of estimators, build on Dempster's rule of combination, Dezert-Smarandache Theory based PCR5 fusion rules and fuzzy-based fusion rule; *(ii)* different types of applied t-conorm and t-norm functions within TCN fusion rule. The algorithm's steps are:

1. Initialization at $k = 0$ :

    - select the frame $\Theta = \{\theta_1,...,\theta_M\}$

    - set the a priori $m^-(\theta_1 \cup ... \cup \theta_M) = 1$. (Full ignorance)

2. Generation of the current measurement from the classifier declaration $T_d(k)$ :

$$m_{obs}(T_d(k)) = c_{T_d(k)T_d(k)}\qquad(15)$$

The unassigned mass is committed to the total ignorance $m_{obs}(\theta_1 \cup ... \cup \theta_M) = 1 - m_{obs}(T_d(k))$.

3. Obtain the updated (estimated) bba $m(.) = [m^- \oplus m_{obs}](.)$ on the base of particular fusion rules: DST, DSmT-PCR5, TCN with different t-conorm *(sum, max)* and t-norm functions, described in equations (6), (7), (8).

4. Estimation of True Target Type by taking the hypothesis having the maximum of belief (or eventually the maximum Pignistic Probability [8]).

5. Propagation to next time step $k+1$

# 5 Simulation results

In order to evaluate the performances of all considered estimators, based on: Dempster's, DSmT based PCR5, TCN fusion rules with different types of t-conorm and t-norm functions applied, and in order to have a fair comparative analysis, Monte-Carlo simulations on a simple scenario for a 2D Target Type frame, i.e. $\Theta = \{Fighter, C\,arg\,o\}$ is done for a classifier with the following confusion matrix:

$$C = \begin{bmatrix} 0.9 & 0.1 \\ 0.1 & 0.9 \end{bmatrix}$$

Here we assume there are two closely spaced targets: Cargo and Fighter. Due to circumstances, attribute measurements received are predominately from one or another and both targets generates actually one single (unresolved kinematics) track. To simulate such scenario, a true Target Type sequence over 100 scans was generated according figure 1 below. The sequence starts with the observation of a Cargo type and then the observation of the target type switches two times onto Fighter type during different time duration. At each time step *k* the decision $T_d(k)$ is randomly generated according to the corresponding row of the confusion matrix of the classifier given the true target type (known in simulations). Then the algorithm presented in the previous section is applied. The simulation consists of 10000 Monte-Carlo runs. The computed averaged performances are shown on the figures 2 and 3 on the base of estimated belief masses obtained by the trackers based on Demspter's rule, DSmT based PCR5 rule, and TCN fusion rule with the following t-conorm and t-norm functions: ***(1) TCN with t-corm=max and t-norm=bounded product; (2)TCN with t-corm=max and t-norm=min; (3) TCN with t-corm=sum and t-norm=min***. It is evident that the target tracker based on Dempster's rule ([10]) is unable to track properly and quickly the changes of target type. Such a very long latency delay is due to the too long integration time necessary to the Demspter's rule for recovering the true belief estimation. PCR5 rule can quickly detect the type changes ([10]) and properly re-estimates the belief masses contrariwise to Dempster's rule. Our goal here is to evaluate the performance of TCN fusion rule with the analyzed above combinations between applied t-conorm and t-norm functions. In general it is obvious that TCN fusion rule shows a stable, quite proper and effective behavior, following the performance of PCR5 rule. TCN fusion rule with *t-corm=max* and *t-norm=bounded product* reacts and adopts better than TCN with t-*corm=sum* and *t-norm=min*, followed by TCN with *t-corm=max and t-norm=min*. As a whole, TCN rule is more cautious in the process of re-estimating the belief masses in comparison to PCR5. It provides symmetric type estimation contrariwise to Dempster's.

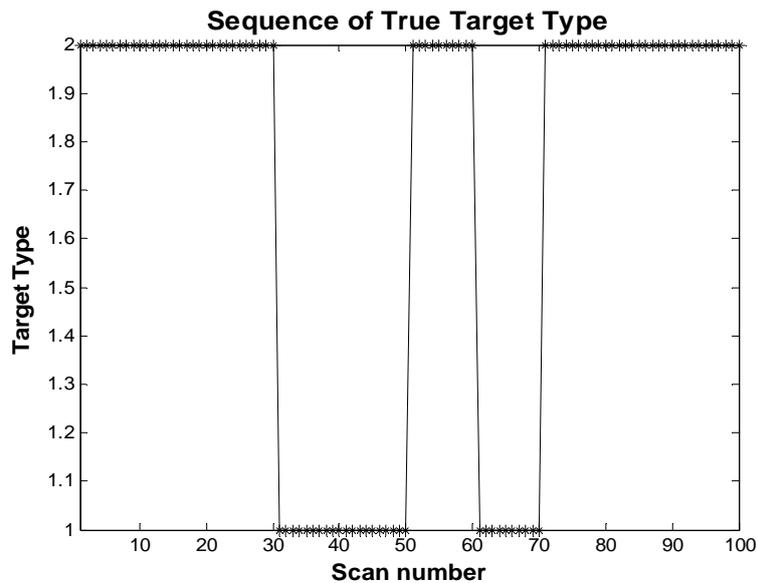

Fig.1 Sequence of true target type (Ground truth)

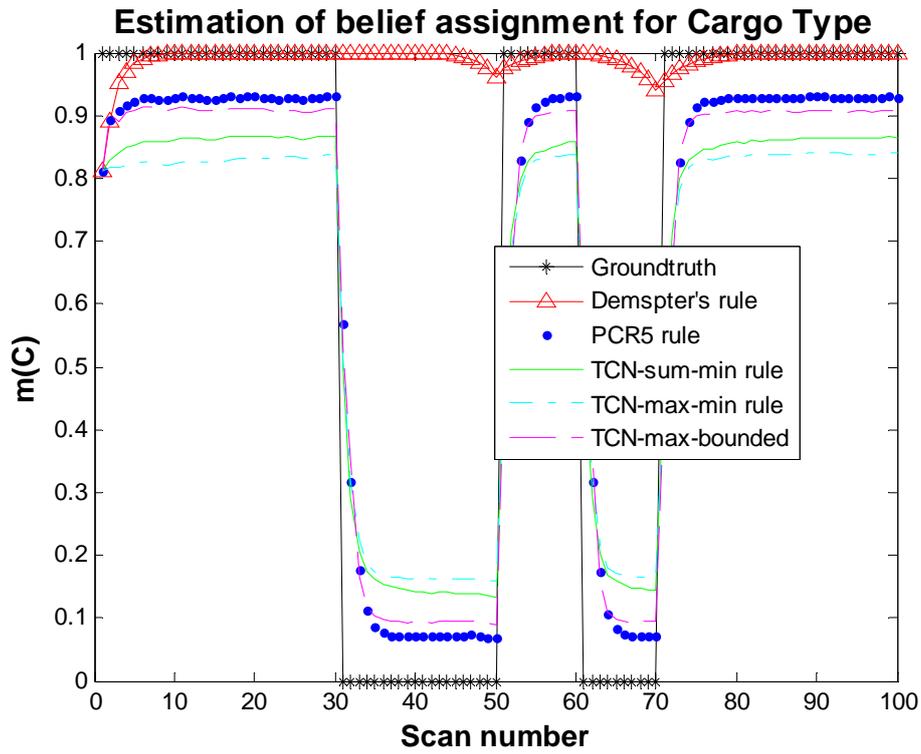

Fig.2 Estimation of belief assignment for Cargo type

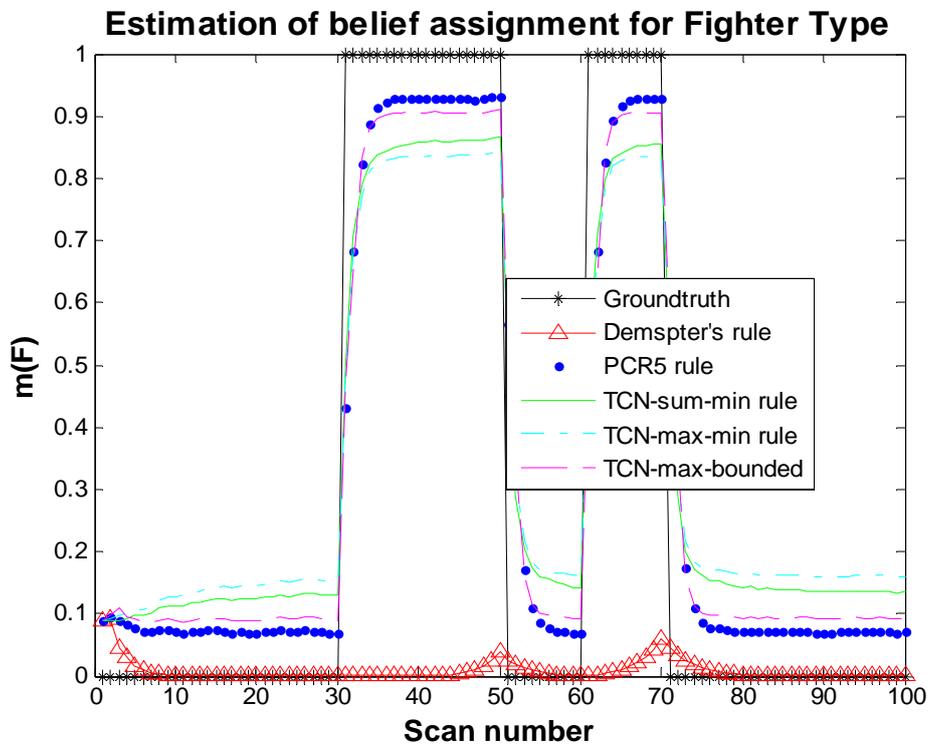

Fig.3 Estimation of belief assignment for Fighter type

# 6 Conclusions

The behavior of three combinational rules for temporal/sequential attribute data fusion for target type estimation has been estimated and analyzed. The comparative analysis is based on: Dempster's fusion rule proposed in Dempster-Shafer Theory; Proportional Conflict Redistribution rule no. 5 (PCR5), proposed in Dezert-Smarandache Theory and on the T-Conorm-Norm alternative class fusion rule, connecting the combination rules for information fusion with particular fuzzy operators, focusing on the t-norm based conjunctive rule as an analog of the ordinary conjunctive rule and t-conorm based disjunctive rule as an analog of the ordinary disjunctive rule. The way how different t-conorms and t-norms functions within TCN fusion rule influence over the target type estimation performance were studied and estimated on the base of simple scenario and Monte-Carlo simulations. In general the different t-conorm, t-norms, available in fuzzy set/logic theory provide us with richness of possible choices to be used in applied here TCN fusion rule. In engineering applications however, the most suitable t-norms are the product or the minimum, because of simplicity of computation and in order to preserve the principle of cause and effect. The attractive features of the new rule could be defined as: very easy to implement, satisfying the impact of neutral Vacuous Belief Assignment; commutative, convergent to idempotence, reflects majority opinion, assures adequate data processing in case of partial and total conflict between the information granules. It is appropriate for the needs of temporal fusion. The general drawback of this rule is related to the lack of associativity, which is not a main issue in temporal data fusion. The target type tracker built on TCN fusion rule shows a stable, quite proper and effective behavior, following the performance of PCR5 rule and overcoming the limitations and drawbacks of DST based tracker.

**Acknowledgement:** This work is founded and supported by the Bulgarian National Science Fund- grant MI-1506/05.

Albena Tchamova
Bulgarian Academy of Sciences,
Institute for Parallel Processing,
Acad. G. Bonchev' Str., Bl. 25-A,
1113 Sofia,
BULGARIA
E-mail: : tchamova@bas.bg

Jean Dezert
ONERA,
29 Av. de la Division Leclerc,
92320 Chatillon,
FRANCE
E-mail: Jean.Dezert@onera.fr

Florentin Smarandache ,
University of New Mexico,
Chair of Department of Mathematics,
Gallup, NM 87301,
USA
E-mail: smarand@unm.edu